# Towards the Feasibility Analysis and Additive Manufacturing of a Novel Flexible Pedicle Screw for Spinal Fixation Procedures

Yash Kulkarni[1], Susheela Sharma[1], Jared Allison[1], Jordan Amadio[2], Maryam Tilton[1], and Farshid Alambeigi[1]

[1]Department of Mechanical Engineering, The University of Texas at Austin, TX 78712
[2]Department of Neurosurgery, The University of Texas Dell Medical School, TX 78712

## Abstract

In this paper, we explore the feasibility of developing a novel flexible pedicle screw (FPS) for enhanced spinal fixation of osteoporotic vertebrae. Vital for spinal fracture treatment, pedicle screws have been around since the early 20th century and have undergone multiple iterations to enhance internal spinal fixation. However, spinal fixation treatments tend to be problematic for osteoporotic patients due to multiple inopportune variables. The inherent rigid nature of the pedicle screw, along with the forced linear trajectory of the screw path, frequently leads to the placement of these screws in highly osteoporotic regions of the bone. This results in eventual screw slippage and causing neurological and respiratory problems for the patient. To address this problem, we focus on developing a novel FPS that is structurally capable of safely bending to fit curved trajectories drilled by a steerable drilling robot and bypass highly osteoporotic regions of the vertebral body. Afterwards, we simulate its morphability capabilities using finite element analysis (FEA). We then additively manufacture the FPS using stainless steel (SS) 316L alloy through direct metal laser sintering (DMLS). Finally, the fabricated FPS is experimentally evaluated for its bending performance and compared with the FEA results for verification. Results demonstrate the feasibility of additive manufacturing of FPS using DMLS approach and agreement of the developed FEA with the experiments.

## Introduction

Bones in a human body consist of two regions: an outer cortical bone layer along with an inner cancellous bone layer [1]. Due to the structure and function of the outer cortical bone layer, it is stronger and stiffer compared to the inner cancellous bone layer which has more of a porous composition [2][3]. While these two healthy bone components work together to support your body through any daily challenges [4], they tend to degrade with age leading to a decrease in bone mineral density (BMD) [5], therefore leading to a decrease in overall strength of the bone layers. The mass and strength of cancellous bone is further decreased when affected by osteoporosis [6]. Osteoporosis is a common health concern defined by a 2.5 standard deviation decrease in BMD compared to the healthy population mean [7] [8]. Osteoporosis is a common disease in the geriatric population with the number of expected bone fractures due to osteoporosis expected to increase to 3.2 million by 2040 [5][9].

Vertebral compression fractures are the most common type of osteoporotic bone fractures with more than 1.4 million global occurrences in patients over the age of 50 [10]. The most common method for fixing this osteoporotic fracture is through spinal fixation (SF) surgery. As shown in Fig. 1, this surgery fuses two vertebral bodies together in a two-step process: (i) drilling a straight corridor through pedicle region of the vertebral body using a rigid drilling instrument and (ii) inserting a rigid pedicle screw (RPS) through the pedicle corridor of the vertebral body and fixating it in the cancellous bone regions. The two vertebral bodies are then locked together through locking rods, eventually returning stability back to the spine. While this procedure has become the gold standard, pedicle screw loosening and pullout remains a critical shortcoming of the current process with RPS fixation failing to provide sufficient fixation stability in osteoporotic bone (i.e., BMD below 80 mg/cm$^3$) [11][12]. Furthermore, the rate of pedicle screw loosening and pullout have reported incidences between 22-50% ([13],[14]) even in healthy BMD scenario. A wide variety of innovations have been explored to prevent the aforementioned screw loosening and pullout issue ranging from changing the geometry of the RPS [15][16] to introducing robotic techniques [17]. Nevertheless, these approaches marginally assist in preventing the aforementioned loosening and pullout problems of pedicle screws. These issues are related to the complex morphology of the vertebra along with the locations of the spinal cord and nerves. The primary cause of pedicle screw loosening and failure can be attributed to the lack of dexterity associated with current rigid drilling instruments and RPS [18][19][20][21]. Figure 1A illustrates how this lack of dexterity restricts the surgeon to place RPS along a linear trajectory, forcing fixation in low BMD regions of the vertebra resulting in complications such a screw loosening and pullout.

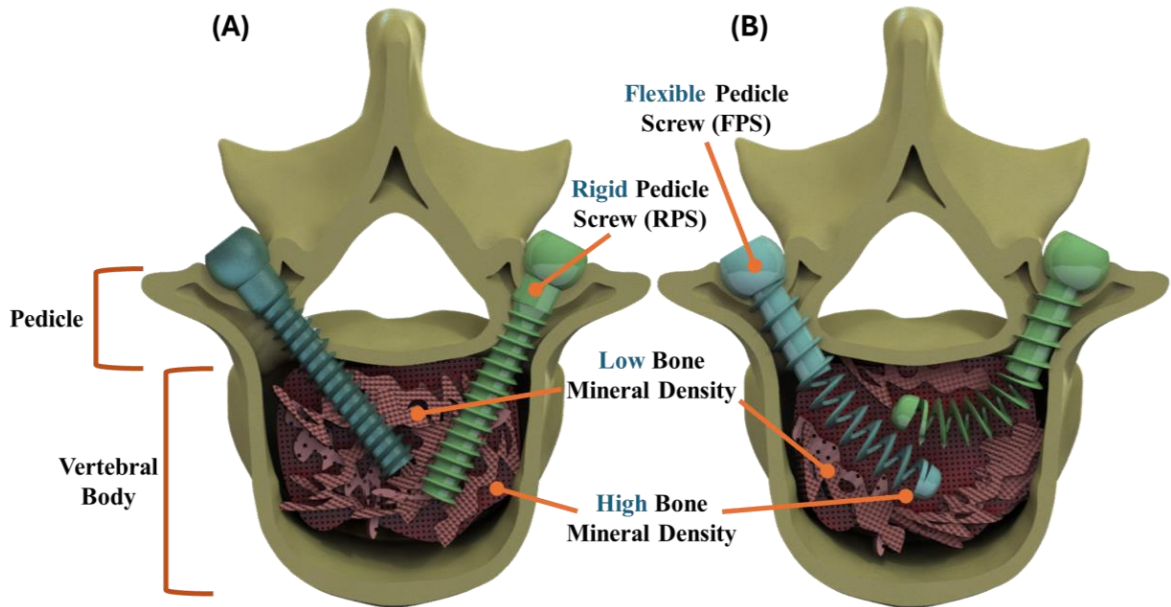

Fig. 1: Conceptual illustration of a pedicle screw fixation process. Left: RPS fixed into areas of low BMD. Right: FPS avoiding areas of low BMD and fixating in high BMD areas.

To address the aforementioned limitations of existing SF with RPSs, this paper studies the feasibility of the flexible pedicle screw (FPS) design and additive manufacturing of a prototype version of a FPS. Figure 1B highlights the potential capabilities of the proposed FPS to avoid those low BMD regions of the vertebral body. First, we will design the FPS and highlight critical design features that are vital in allowing the FPS to safely and reliably follow a curved trajectory created by our previously introduced concentric tube-steerable drilling robot (CT-SDR) [20][21], shown in Fig 2. Then, we will develop a finite element analysis (FEA) model to investigate the morphability capabilities of the FPS and then fabricate the FPS in stainless steel (SS) 316L material using the direct metal laser sintering (DMLS) process. Finally, we will verify the FPS bending capabilities by comparing the FEA simulation and experimental studies.

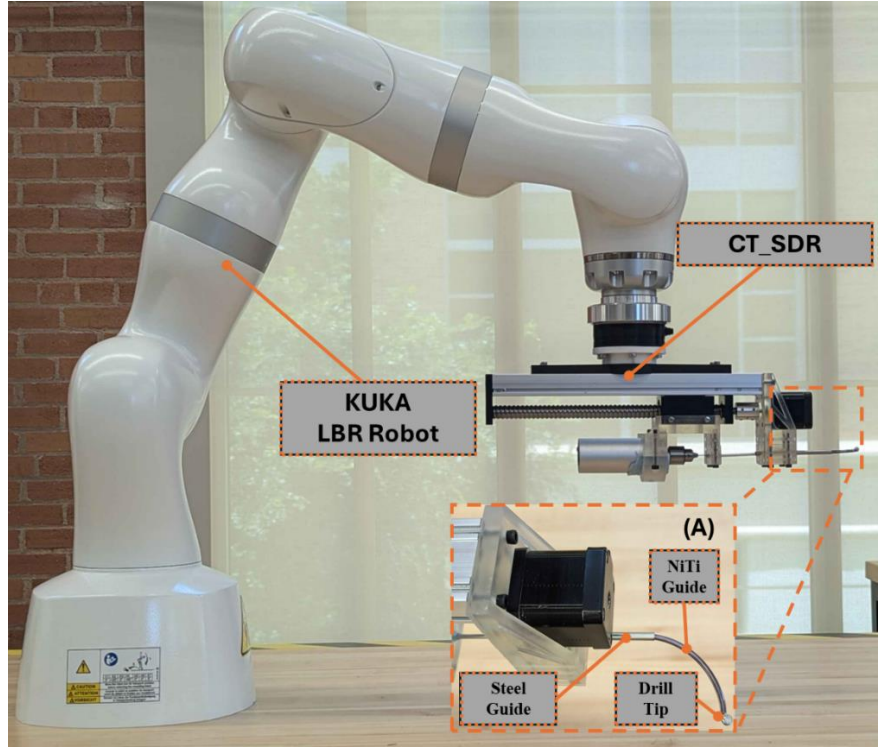

Fig. 2: Set-up illustrating the Kuka robot and the CT-SDR system together. (A): A detailed view into the components of the CT-SDR including the steel guide, nitinol guide, and drill tip.

## Materials and Methods

Driven by the key requirement of the FPS safely bending to fixate in all areas of the vertebral body while remaining structurally strong, the conventional RPS design must be updated as shown in Fig. 3. The RPS design consists of a rigid, rod-like shaft that only allows for fixation in a linear trajectory. Therefore, the FPS requires critical geometric design changes to safely morph and fit the curved trajectories created by the CT-SDR. Building on the previous work by Alambeigi et al. [18] [19] and Kulkarni et al [22], we propose the following features:

**Safe Morphing Requirements:** The safety of the FPS relies on its ability to bend and follow curvilinear trajectories created by the CT-SDR without breaching the surrounding bone while still ensuring stability within the vertebral body. Figure 3 introduces the four unique design features of the FPS to ensure the aforementioned requirements: (i) a semi-flexible region, (ii) a semi-rigid region, (iii) a rounded tip, and (iv) a cannulated region (goes through the center of the FPS).

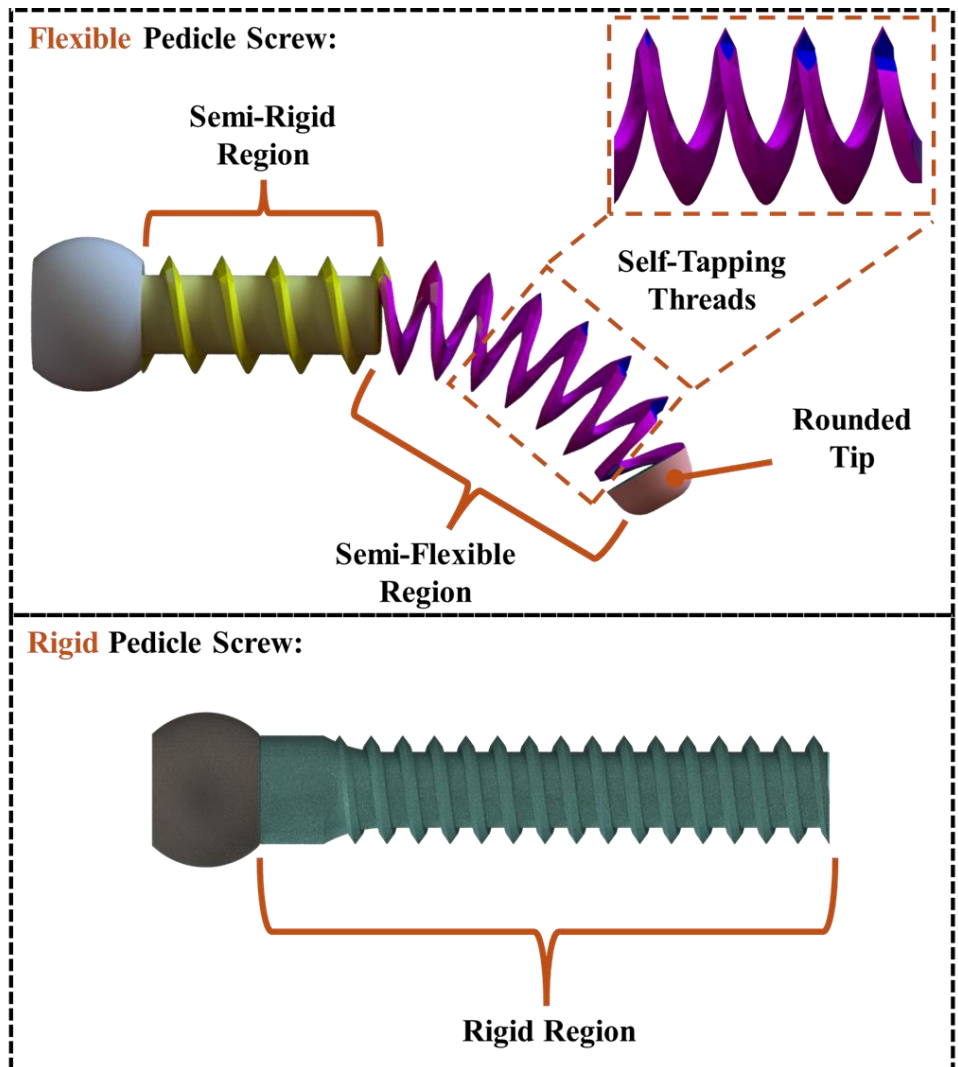

Fig. 3: Comparative conceptual designs of the novel FPS (top) and a conventional RPS (bottom). The FPS is comprised of five unique design components: the semi-rigid region, the semi-flexible region, self-tapping threads, rounded tip, and a cannulated region. The RPS only has a singular rigid region.

The *semi-rigid and semi-flexible* regions work together to ensure the success of the FPS. The semi-flexible region enables the FPS to bypass low BMD areas in the vertebral body, which are prone to pedicle screw pullout failures, and fixate in high BMD regions. Meanwhile, the semi-rigid region ensures stability by working in tandem with the rigid pedicle area of the vertebra. Additionally, the gap in the flexible region promotes bone growth, potentially enhancing long-term implant security through better bone-screw integration, thereby reducing the risk of screw pullout. Without loss of generality and as proof of concept, an L3 vertebra's shape and geometry [23] along with the capabilities of the CT-SDR were used to drive the dimensions of the FPS. With CT-SDR [20][21] creating an average pilot hole size of 8 mm in diameter, the FPS was designed with a 9.0 mm outer diameter (OD) with a 6.0 mm inner diameter (ID). Furthermore, this FPS had a 4.0 mm pitch with a 3.0 mm thread height ($T_H$) and a 3.0 mm inner diameter within the flexible region of the FPS ($F_{ID}$). This screw was designed with a 30.8 mm flexible region ($L_f$) and an 18.0 mm rigid region ($L_r$). These parameters are highlighted on Fig. 4. Of note, these dimensions can readily be modified and redesigned for a specific level of vertebra and based on a patient-specific FE analysis [24].

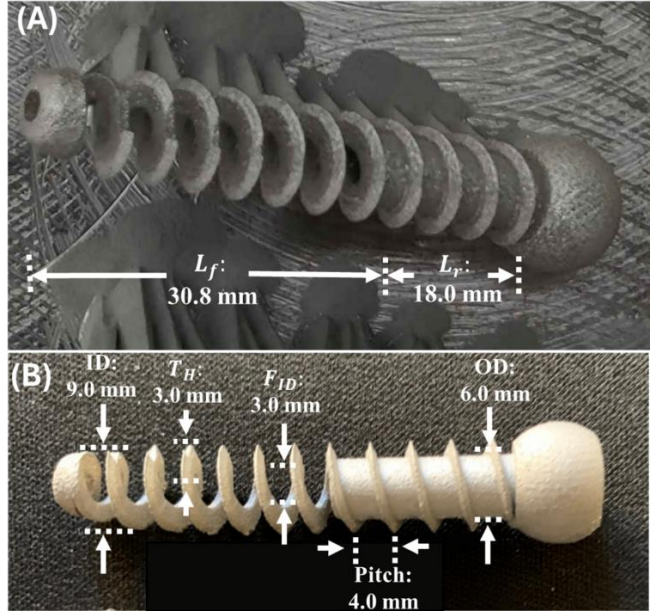

Fig. 4: Metal 3D printed flexible pedicle screw. (A) The fabricated FPS on the titanium build plate before removal. (B) The two different sized pedicle screws after post-processing.

The *rounded tip* ensures that the FPS safely follows the curved trajectory created by the CT-SDR.

Also, the *cannulated* region of the FPS allows for potential increased stability in the vertebral body by providing a path for bone cement augmentation, as needed. This potential path for bone cement injection increases the possibility of enhanced fixation. For the designed FPS, we have considered a cannulated region with 3 mm.

**FPS FE Model:** To accurately design the FPS, we also created an FEA model in *ANSYS Workbench* (Canonsburg, PA, USA) based on the aforementioned design parameters. In our simulations, we solely focused on modeling bending behavior of the FPS to analyze its morphability feature (defined as the ability to follow a predrilled curved trajectory by the steerable drilling robot, as shown in Fig. 1B). Therefore, for these studies, as shown in Fig. 5, the rigid part of the screw was fixed while a 6 mm displacement condition being applied to the rounded tip of the FPS. Because of the compliant nature of the FPS and to obtain accurate FE results, a large deflection condition was considered in the software.

Using this model and due to the orthotropic material properties of the additive manufacturing process, we performed a sensitivity analysis by varying Young's moduli while keeping the rest of the material parameters and boundary conditions constant. The stainless-steel FPS was assigned a Young's modulus ranging from 155-185 GPa in the XY direction [25],[26] with the Young's modulus ranging from 150-180 GPa in the Z direction. The Z directions Young's modulus was modeled to be 5 GPa lower than the XY direction based on the difference seen on the manufacturer's datasheet provided (185 GPa in XY direction with 180 in Z direction [25]). In order to ensure the sensitivity analysis solely focused on the Young's modulus, a constant Poisson ratio of 0.3 and a calculated shear modulus of 59 GPa, based on the formula for shear modulus of orthotropic materials at 155/150 GPa [27], was used for all models along the XY and Z direction. Of note, after the FPS was fabricated, this analysis was then compared with our experimental studies (discussed in the next section) to further understand our FE model. Figure 6 shows the obtained simulation forces needed to displace the tip of FPS to the maximum deflection of 6 mm for the different considered Young's modulus.

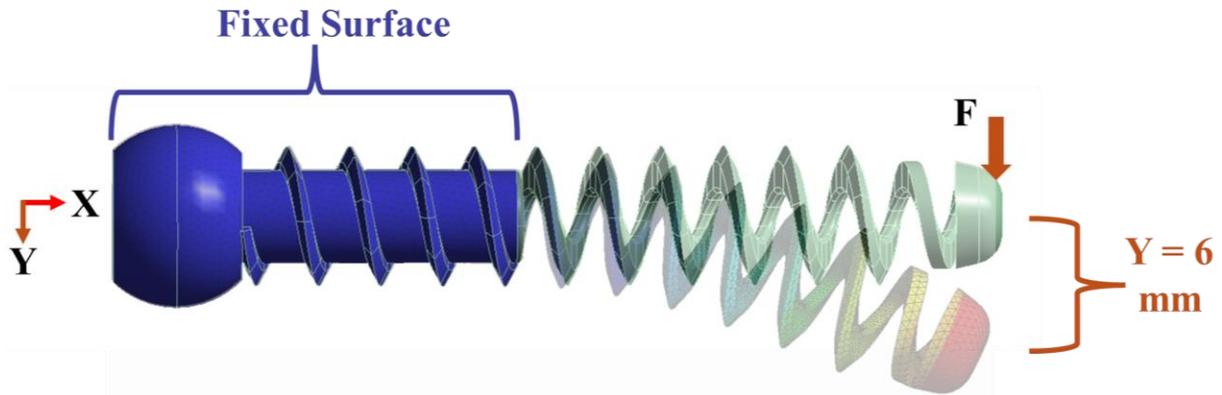

Fig. 5: Finite element model of the FPS simulating its bending capabilities with the boundary conditions and displacement labeled

**Fabrication Requirements:** To ensure the FPS's complex geometry can be fabricated reliably, the Direct Metal Laser Sintering (DMLS) process was utilized using the EOS M280 machine. The FPS was fabricated from SS 316L granulates sourced from EOS (Krailling, Germany). The FPS were sintered with a layer thickness of 20 microns. The prints were built diagonally with an approximate angle of 35 degrees with respect to the printing bed with multiple stainless-steel support being used. These supports being critical in preventing any residual stress effects warping the thread formation during the additive manufacturing process. The supports were manually removed after the manufacturing process and then smoothened using a sanding tool.

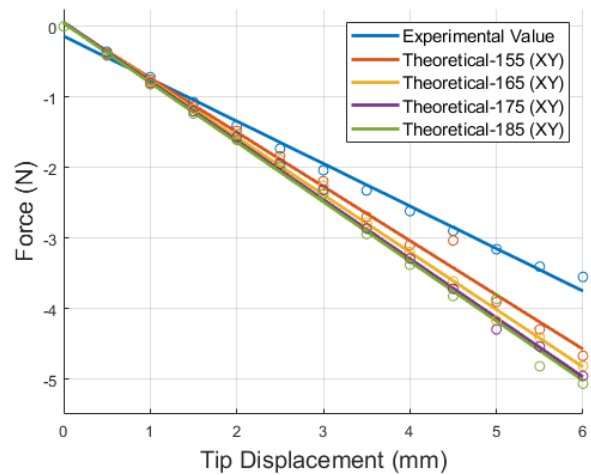

Fig. 6: The graph illustrates the experimental force-displacement results overlaid upon multiple ranging theoretical FEA simulations for both 9mm screws screw SS 316L screw.

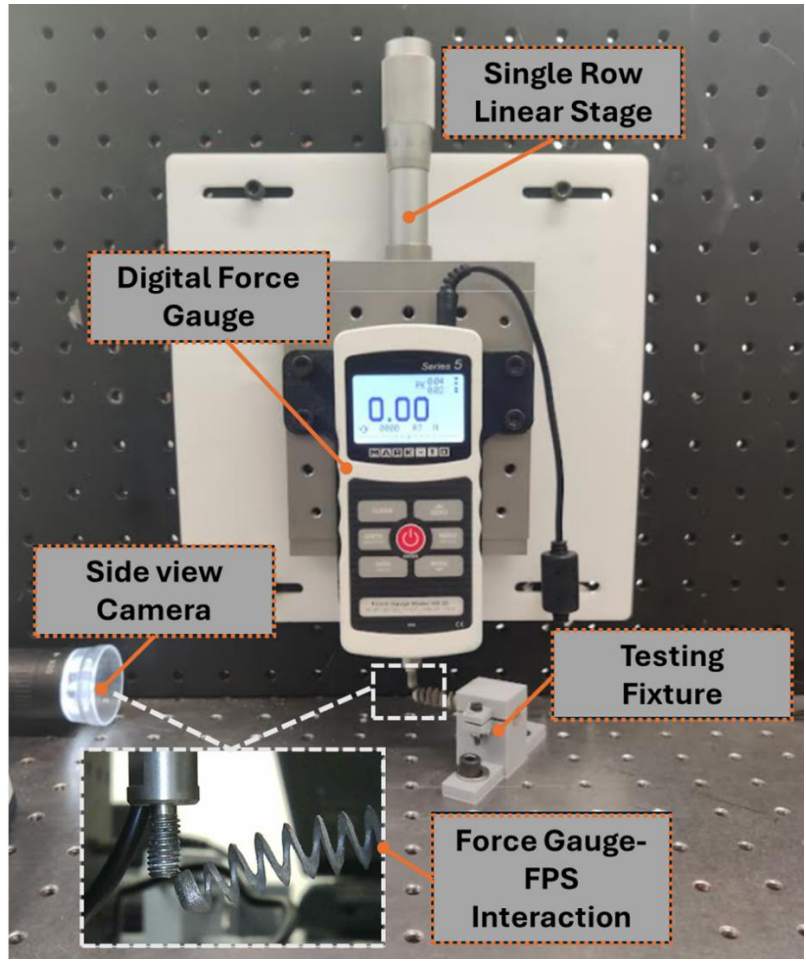

Fig. 7: Experimental set-up utilized to conduct force-displacement testing.

**Experiment Setup and Procedure**

Due to the importance of FEA in the development of the FPS, an in-depth experimental study validating the FEA results against experimental results was conducted. The overall goal was to measure the efficacy and capability of the simulation to match real world results regarding the morphability feature and bending ability of FPS by measuring its distal end deformation under various loads. To this end and to gather accurate and continuous force-displacement data, the experimental set-up shown in Fig. 7 was created. The testing fixture was created using Solidworks software (Dassault Systèmes) and was printed using Raise3D E2 printer (Costa Mesa, California) to firmly hold the SS 316L FPS during the experiments. A digital force gauge Mark-10 Series 5, Mark Ten) with a resolution of 0.02 N was coupled with a single-row linear stage (M-UMR12.40) with 1 $\mu$m precision to accurately deflect the tip of the FPS down to measure the reaction force. The entire system was secured to an optical table to prevent slippage and other errors from being introduced to the results. During the experiment, the tip of the force gauge was brought slowly down via the linear stage until it barely touched the tip of the screw. The side camera was utilized during this process to detect when the two tips met. After the tips met, the force gauge was zeroed out, and the linear stage was rotated in 0.5 mm increments up to 6 mm, therefore imposing a 6 mm tip displacement on the screw tip. During the process, we obtained the interaction force measured by the force gauge corresponding to the tip deflection of the screw. The force gauge was then moved upward until there was no visible interaction between the screw tip and the force gauge tip. The experiment was then repeated three times, with the results being averaged out. The obtained experimental force-displacement data was then used to verify the developed FE model. Figure 6 compares the obtained experimental forces needed to displace the tips of the fabricated 9 mm SS 316L FPS with the performed simulation studies.

TABLE I: Mean absolute error (MAE) and root mean square error (RMSE) measured between the theoretical predictions and experimental results along with the maximum, minimum, and standard deviation between the difference of the experimental results and theoretical predictions

| Flexible Pedicle Screw Model | Young's Modulus (XY)/(Z) (GPa) | MAE (N) | RMSE (N) | Maximum Difference (N) | Minimum Difference (N) | Standard Deviation (N) |
|---|---|---|---|---|---|---|
| 9 mm | 155/150 | 0.35 | 0.50 | 1.12 | 0.020 | 0.38 |
|  | 165/160 | 0.48 | 0.63 | 1.26 | 0.032 | 0.43 |
|  | 175/170 | 0.55 | 0.73 | 1.40 | 0.043 | 0.48 |
|  | 185/180 | 0.59 | 0.76 | 1.51 | 0.053 | 0.50 |

## Results and Discussion

Figure 6 shows the forces required to displace the tip of the FPS until it reaches a maximum displacement of 6 mm in both the FE simulation and the experimental study. For the experimental study, the maximum force of 3.55 N is required to reach this maximum displacement. The FE simulation is able to match this closely at 4.67 N at a Young's Modulus of 155/150 GPa with a standard deviation of 0.38 N between the FE simulation and experimental results. The Young's moduli values of 165/160 GPa, 175/170 GPA, and 185/180 GPa required a higher maximum force of 4.82 N, 5.00 N and 5.07 N to reach the same displacement, respectively. These values had a low standard deviation of 0.43 N, 0.48 N, and 0.50 N between the FE simulation and experimental study, respectively. As predicted, the higher Young's modulus simulations required more force to morph the FPS to the same tip displacement.

Table 1 summarizes the results discussed above along with the maximum, minimum, and standard deviation between the FE prediction and experimental results. As can be observed the FE simulation with a Young's modulus of 155/150 GPa resulted in the best mean absolute error (MAE) of 0.3503 N and a root mean squared error (RMSE) of 0.5033 N as compared with the experimental results. Also, as we increase the Young's modulus, the difference between the FE analysis and experimental results increases. This analysis indicates that designing additively manufactured FPSs in FEA with the Young's modulus 155/150 GPa would output a more realistic result as compared with the other considered values. Of note, this obtained Young's modulus is within the accepted range reported in the literature [25],[26]. It is also worth noting that the discrepancies between the FE model and experimental results could have been reduced by focusing on an entire model tuning rather than changing the theoretical Young's modulus solely. Other reasons for discrepancy's witness could be due to imperfection of the fabrication and post processing procedure as compared with the designed CAD model.

## Conclusion:

In this paper, we presented the feasibility of the design and additive manufacturing of a novel flexible implant for spinal fixation procedures. The proposed FPS has the potential of overcoming the current limitations of a rigid pedicle screw by minimizing the risk of implant pullout and loosening. Our results demonstrate (i) the feasibility of metal additive manufacturing the proposed flexible implant with stainless-steel and (ii) a maximum MAE error of 0.59 N and RMSE error of 0.76 N between the developed FE model and experimental results, proving its reasonable accuracy to be used for the design and modeling of different FPSs.

Despite the promising results, in the future we will improve our FE model to reduce the obtained errors and optimize the design of the FPS. Using this model, we will design and fabricate different types of FPSs with various sizes, parameters, and materials (e.g., titanium). We will also evaluate the FPS's morphability capabilities, fixation and pullout strength in both Sawbone and cadaveric specimen to thoroughly evaluate its fixation capabilities.

## Acknowledgement:

This work is supported by the National Institute Of Biomedical Imaging and Bioengineering of the National Institutes of Health under Award Number R21EB030796.